%% file: ecai2020_division.tex
\documentclass{ecai}
\usepackage{times}
\usepackage{graphicx}
\usepackage{latexsym}

\usepackage{soul}
\usepackage{url}
\usepackage[hidelinks]{hyperref}
\usepackage[utf8]{inputenc}
\usepackage[small]{caption}
\usepackage{graphicx}
\usepackage{amsmath}
\usepackage{booktabs}
\usepackage{algorithm}
\usepackage{algorithmicx}
\usepackage{algpseudocode}
\usepackage{longtable} 
\usepackage{amsmath}
\usepackage{amssymb}

\usepackage{url}
\usepackage[colorinlistoftodos]{todonotes}
\usepackage{paralist}
\usepackage{xspace}
\usepackage{subcaption}
\usepackage{lipsum}
\usepackage{bm}
\usepackage{paralist}

 \usepackage{multicol} 
 \usepackage{multirow} 

\usepackage{lipsum}

\usepackage[shortlabels]{enumitem}
\setlist[enumerate]{nosep}

\usepackage{xspace}

\input{macros.tex}

\renewcommand{\and}{{\hspace{0.07em} \rm and \hspace{0.05em}}}

\ecaisubmission   

\newcommand\blfootnote[1]{%
  \begingroup
  \renewcommand\thefootnote{}\footnote{#1}%
  \addtocounter{footnote}{0}%
  \endgroup
}

\begin{document}

\title{Dividing the Ontology Alignment Task\\with Semantic Embeddings and Logic-based Modules\blfootnote{Accepted to the 24th European Conference on Artificial Intelligence~(ECAI~2020)}}

\author{Ernesto Jim\'enez-Ruiz\institute{City, University of London, UK, email: ernesto.jimenez-ruiz@city.ac.uk; and Department of Informatics, University of Oslo, Norway} \and Asan Agibetov\institute{Section for Artificial Intelligence and Decision Support, Medical University of Vienna,
Vienna, Austria} \and Jiaoyan Chen\institute{Department of Computer Science, University of Oxford, UK} \and Matthias Samwald$^{3}$\and Valerie Cross\institute{Miami University, Oxford, OH 45056, United States} }

\maketitle
\bibliographystyle{ecai}

\begin{abstract}
 Large ontologies still pose serious challenges to state-of-the-art ontology
alignment systems. In~this paper we present an approach that combines a neural
embedding model and logic-based modules to 
accurately
divide an 
input 
ontology
matching task into smaller and more tractable matching (sub)tasks. 
We have conducted a comprehensive evaluation using the datasets of the Ontology Alignment Evaluation Initiative. 
The results are encouraging and suggest that the proposed method is adequate in
practice and can be integrated within the workflow of systems unable to cope
with very large ontologies.
\end{abstract}

\input{introduction}

\input{background}

\input{methods}

\input{evaluation}

\input{related}

\input{discussion}


\ack 
This work was supported by the SIRIUS Centre for Scalable Data Access (Norges forskningsr{\aa}d), the AIDA project (Alan Turing Institute), Samsung Research UK, Siemens AG, and the EPSRC projects AnaLOG, OASIS and UK FIRES. We would also like to thank the anonymous reviewers that helped us improve~this~work.


\bibliographystyle{ecai}
\bibliography{ecai}
\end{document}

%% file: macros.tex
\newtheorem{definition}{Definition}

\newcommand{\ie}{\textit{i}.\textit{e}.,\xspace}
\newcommand{\eg}{\textit{e}.\textit{g}.,\xspace}

\newcommand{\ontoOne}{\ensuremath{\O_1}}
\newcommand{\ontoTwo}{\ensuremath{\O_2}}

\newcommand{\FMA}{\ensuremath{\O_\text{FMA}}}
\newcommand{\NCI}{\ensuremath{\O_\text{NCI}}}

\newcommand{\ontoOneP}[1]{\ensuremath{\ontoOne}^{#1}}
\newcommand{\ontoTwoP}[1]{\ensuremath{\ontoTwo}^{#1}}

\renewcommand{\O}{\ensuremath{\mathcal{O}}\xspace}
\newcommand{\M}{\ensuremath{\mathcal{M}}}
\newcommand{\T}{\ensuremath{\mathcal{T}}}

\newcommand{\D}{\ensuremath{\mathcal{D}}}
\newcommand{\mapping}[4]{\langle #1,\allowbreak #2,\allowbreak #3,\allowbreak #4
\rangle}
\newcommand{\mappingTwo}[2]{\langle #1,\allowbreak #2 \rangle}
\newcommand{\myset}[1]{ \{#1\} }
\newcommand{\MS}{\M^{S}\xspace}
\newcommand{\MRA}{\M^{RA}\xspace}
\newcommand{\MRAFN}{\MRA_{\text{fma-nci}}\xspace}

\newcommand{\MLex}{\M^{\lex}\xspace}
\newcommand{\MT}{\ensuremath{\M\T}\xspace}
\newcommand{\MTLex}{\MT^{\lex}\xspace}
\newcommand{\PMT}{\ensuremath{\D_{\MT}}\xspace}

\newcommand{\MTMO}{\MT^{\M}_{\O_1\text{-}\O_2}\xspace}
\newcommand{\MTRAFN}{\MT^{RA}_{\text{fma-nci}}\xspace}
\newcommand{\MTFN}{\MT_{\text{fma-nci}}\xspace}

\newcommand{\cem}[1]{\mathsf{#1}}
\newcommand{\nsp}{\negthickspace}

\newcommand{\lex}{\textsf{LexI}\xspace}
\newcommand{\starspace}{\textsf{StarSpace}\xspace}

\newcommand{\logicalalgo}[1]{\ifmmode \text{ \textbf{#1} } \else \textbf{#1} \fi}

\newcommand{\funcalgo}[1]{\ensuremath{\mathsf{#1}}}

%% file: introduction.tex
\section{Introduction}

The problem of (semi-)automatically computing an alignment between
independently developed ontologies has been extensively studied in the last
years. As a result, a number of sophisticated ontology alignment systems currently
exist \cite{DBLP:journals/tkde/ShvaikoE13,DBLP:books/daglib/0032976}.\footnote{Ontology matching surveys and approaches: \url{http://ontologymatching.org/}} 
The Ontology
Alignment Evaluation Initiative\footnote{OAEI evaluation campaigns:
\url{http://oaei.ontologymatching.org/}} (OAEI)
\cite{DBLP:conf/semweb/AlgergawyCFFFHH18,DBLP:conf/semweb/AlgergawyFFFHHJ19} has played a key role in the benchmarking of these systems by
facilitating 
their comparison on the same basis and 
the reproducibility of the results. 
The OAEI includes different tracks organised by different research groups.
Each track contains one or more matching tasks involving small-size (\eg
conference), medium-size (\eg anatomy), large (\eg phenotype) or
very large (\eg largebio) ontologies.
Some tracks only involve matching at the terminological level (\eg concepts and properties) while other tracks also expect an alignment at the assertional level (\ie instance data).

Large ontologies still pose serious challenges to ontology alignment systems.
For example, several 
systems participating in the \emph{largebio track} were unable to complete the
largest tasks during the latest OAEI 
campaigns.\footnote{Largebio track: \url{http://www.cs.ox.ac.uk/isg/projects/SEALS/oaei/}}
These systems typically use advanced alignment 
methods and are able to cope with small and medium size
ontologies with competitive results, but fail to complete large tasks in a given time frame or with the available resources 
such as memory.

There have been several efforts in the literature to divide the ontology
alignment task (\eg \cite{Hamdi:2010,hu:2008}).
These approaches, however, have not been successfully 
evaluated with very large ontologies, failing to scale or producing partitions 
of the ontologies leading to information loss \cite{pereira:2017}.
In this paper we propose a novel method to accurately divide the matching task
into several independent, smaller and manageable (sub)tasks, so as to scale
systems that cannot cope with very large ontologies.\footnote{A preliminary version of this work has been published in arXiv \cite{arxiv18_division} and in the Ontology Matching workshop \cite{om18_division}.}
Unlike state-of-the-art approaches, 
our method:
\begin{inparaenum}[\it (i)]
\item 
preserves the coverage of the relevant ontology alignments while keeping
manageable matching subtasks;
\item provides a formal notion of matching subtask and semantic context;
\item uses neural embeddings to compute an accurate division by learning
semantic similarities between words and ontology entities according to the
ontology alignment task at hand;
\item computes self-contained (logical) modules to guarantee the inclusion of the (semantically) relevant information required by an alignment system; and
\item has been successfully evaluated with very large ontologies. 
\end{inparaenum}


%% file: background.tex
\section{Preliminaries}
\label{sec:preliminaries}



A \emph{mapping} (also called \textit{match}) 
between entities\footnote{In this work we accept any input ontology in the OWL~2 language \cite{OWL2}. We refer to (OWL~2) concepts, properties and individuals as entities.} of two ontologies $\O_1$ 
and $\O_2$ 
is typically represented as a 
4-tuple $\mapping{e_1}{e_2}{r}{c}$ where $e_1$ and $e_2$ are entities of $\O_1$ and $\O_2$, 
respectively; $r$ is a semantic relation, typically one of $\myset{\sqsubseteq, \sqsupseteq, \equiv}$; and $c$ is a confidence value, 
usually, a real number within the interval $\left(0,1\right]$. For simplicity, we refer to a mapping 
as a pair $\mappingTwo{e_1}{e_2}$.
%
An ontology \emph{alignment} is a
set of mappings $\M$ between two ontologies $\O_1$ and $\O_2$.

An ontology \emph{matching task} $\MT$ is composed of a pair of ontologies
$\O_1$ (typically called source) and $\O_2$ (typically called target) and possibly an associated \emph{reference alignment} $\MRA$.
The objective of a matching task is to discover an overlapping of $\O_1$ and
$\O_2$ in the form of an alignment $\M$. The \emph{size} or \emph{search space}
of a matching task is typically bound to the size of the 
Cartesian product between the entities of the input ontologies:
$\lvert Sig(\O_1) \rvert \times \lvert Sig(\O_2)\rvert$,
where $Sig(\O)$ denotes the signature (\ie entities) of $\O$ and $\lvert \cdot \lvert$ denotes the size of a set.

An ontology \emph{matching system} is a program
that, given as input a matching task $\MT=\langle \ontoOne, \ontoTwo \rangle$,
generates an ontology alignment $\MS$.\footnote{Typically automatic, although there are systems that also allow human interaction \cite{ker2019}.}
The standard evaluation measures for an alignment $\MS$ are
\emph{precision} (P), \emph{recall} (R) and \emph{f-measure} (F) computed against a reference alignment $\MRA$ as follows:
\begin{small}
\begin{equation}\label{eq:measures}
    P = \frac{\lvert\MS \cap \MRA\rvert}{\lvert\MS\rvert},~
    R = \frac{\lvert\MS \cap \MRA\rvert}{\lvert\MRA\rvert},~
    F = 2 \cdot \frac{P \cdot R}{P + R} 
\end{equation}
\end{small}

\begin{figure*}[ht!]
        \centering
        \includegraphics[width=0.99\textwidth]{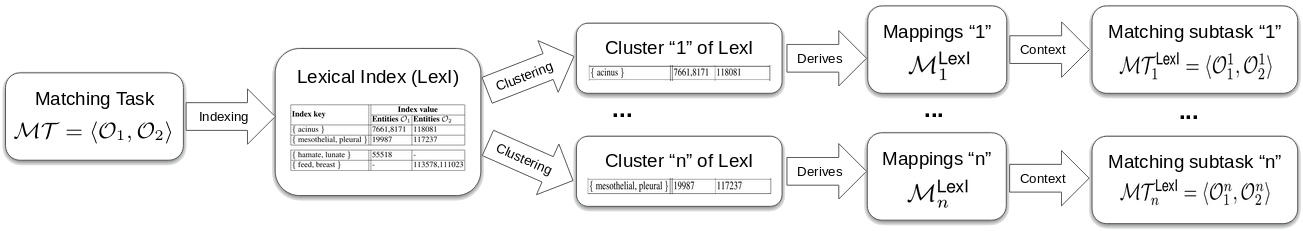}\\[-1ex]
        \caption{Pipeline to divide a given matching task $\MT=\langle \ontoOne, \ontoTwo \rangle$.}%
        \label{fig:pipeline}
    
    \end{figure*}

\subsection{Problem definition and quality measures}
\label{sect:quality}

We denote \emph{division} of an ontology matching task $\MT$, composed by
the ontologies $\O_1$ and $\O_2$, as the process of finding
$n$ matching subtasks $\MT_{i}=\langle \ontoOne^{i}, \ontoTwo^{i} \rangle$ (with
$i$=$1$,\ldots,$n$), where $\ontoOne^{i} \subset \ontoOne$ and $\ontoTwo^{i} \subset
\ontoTwo$. 
%

\medskip
\noindent
\textbf{Size of the division.}
The size of each matching subtask 
is 
smaller than the
original task and thus reduces the search space.
Let $\PMT^{n}=\{\MT_1,\ldots,\MT_n\}$ 
be
the division of a matching task $\MT$ into $n$ subtasks.
The
\emph{size ratio} of the
subtasks $\MT_i$ and $\PMT^{n}$ 
with respect to the original matching task size 
is
computed as follows:

%
\begin{small}
\begin{equation}\label{eq:sizeeratioTask}
    \funcalgo{SizeRatio}(\MT_i, \MT) = \frac{\lvert Sig(\ontoOne^{i}) \rvert
    \times \lvert Sig(\ontoTwo^{i}) \rvert} {\lvert Sig(\O_1) \rvert \times \lvert
    Sig(\O_2)\rvert}
\end{equation}
\end{small}
\begin{small}
\begin{equation}\label{eq:sizeeratio}
    \funcalgo{SizeRatio}(\PMT^{n}, \MT) =
    \sum_{i=1}^{n}\funcalgo{SizeRatio}(\MT_i, \MT)
\end{equation}
\end{small}
The ratio $\funcalgo{SizeRatio}(\MT_i, \MT)$ is
less than
$1.0$ while the aggregation $\sum_{i=1}^{n}\funcalgo{SizeRatio}(\MT_i, \MT)$, being $n$ the number of matching subtasks, 
can be greater than
$1.0$ as matching subtasks depend on the division technique and may overlap.

\medskip
\noindent
\textbf{Alignment coverage.}
The division of the matching task aims at
preserving the target outcomes of the original matching task.
The \emph{coverage} is
calculated with respect to a relevant alignment $\M$, possibly the
reference alignment $\MRA$ 
of the matching task 
if it exists, and 
indicates whether that alignment can still be (potentially) discovered with the matching subtasks.
The formal notion
of coverage is given in Definitions \ref{def:cov1} and \ref{def:cov2}. 

\begin{definition}[Coverage of a matching task]
\label{def:cov1}
Let $\MT=\langle \ontoOne, \ontoTwo \rangle$ be a matching task and $\M$ an
alignment. We say that a mapping $m=\mappingTwo{e_1}{e_2} \in \M$ is covered by
the matching task if $e_1 \in Sig(\O_1)$ and $e_2 \in Sig(\O_2)$. The
coverage of $\MT$ w.r.t. $\M$ (denoted as $\funcalgo{Coverage}(\MT, \M)$) 
represents the set of mappings 
$\M' \subseteq \M$ 
covered~by~$\MT$.
\end{definition}

\begin{definition}[Coverage of the matching task division]
\label{def:cov2}
Let $\PMT^{n}=\{\MT_1,\ldots,\MT_n\}$ be the result of dividing a matching
task $\MT$ and $\M$ an alignment. We say that a mapping $m\in
\M$ is covered by $\PMT^{n}$ if $m$ is at least covered by one of the matching
subtask $\MT_i$ (with $i$=$1$,\ldots,$n$) as in Definition \ref{def:cov1}.
The coverage of $\PMT^{n}$ w.r.t. $\M$ (denoted as $\funcalgo{Coverage}(\PMT^{n}, \M)$) 
represents the set of mappings $\M' \subseteq \M$ covered by $\PMT^{n}$.
The coverage is 
given as a ratio with respect to the (covered)
alignment:
\begin{small}
\begin{equation}\label{eq:covratio}
    \funcalgo{CoverageRatio}(\PMT^{n}, \M) =
    \frac{\lvert\funcalgo{Coverage}(\PMT^{n}, \M)\rvert} {\lvert \M \rvert}
\end{equation}
\end{small}
\end{definition}

%% file: methods.tex
\section{Methods}
\label{sec:methods}

In this section we present our approach to compute a division 
$\PMT^{n}=\{\MT_1,\ldots,\MT_n\}$ given a matching task $\MT=\langle \ontoOne, \ontoTwo \rangle$ and the number of target subtasks $n$.
We rely on locality ontology modules to extract 
self-contained modules of the input ontologies. The module extraction and task division 
is tailored to the ontology alignment task at hand 
by embedding the contextual semantics of a (combined) inverted index of the ontologies in the matching~task.

Figure \ref{fig:pipeline} shows an overview of our approach. 
\textit{(i)} The ontologies $\ontoOne$ and $\ontoTwo$ are indexed using the lexical index \lex (see Section \ref{sec:lex}); \textit{(ii)} \lex is divided into clusters based on the semantic embeddings of its entries (see Section \ref{sec:clustering}); \textit{(iii)} entries in those clusters derive potential mapping sets (see Section \ref{sec:cmt}); and \textit{(iv)} the context of these mapping sets lead to matching subtasks (see Sections \ref{sec:locality} and \ref{sec:cmt}). Next, we elaborate on the methods behind these steps.



\subsection{Locality modules and context}
\label{sec:locality}


Logic-based module extraction techniques compute ontology fragments that capture
the meaning of an input signature 
(\eg set of entities) 
with respect to a given
ontology. That is, a module contains the context 
(\ie sets of \emph{semantically related} entities) of the input signature.
In this paper we rely on bottom-locality modules
\cite{DBLP:journals/jair/GrauHKS08,DBLP:conf/esws/Jimenez-RuizGSSL08}, which will be referred to as locality-modules or simply as modules. These modules include the ontology axioms required to describe the entities in the signature.
Locality-modules compute self-contained ontologies and are tailored to tasks that require reusing a fragment of an ontology. Please refer to \cite{DBLP:journals/jair/GrauHKS08,DBLP:conf/esws/Jimenez-RuizGSSL08} for further details.


Locality-modules play an key role in our approach
as they provide the context 
for the entities in a given mapping or set of mappings as formally
presented in Definition~\ref{def:context1}.

\begin{definition}[Context of a mapping and an alignment]
\label{def:context1}
Let $m=\mappingTwo{e_1}{e_2}$  be a mapping
between two ontologies $\O_1$ and $\O_2$. We define the context of $m$ (denoted
as $\funcalgo{Context}(m, \O_1, \O_2)$) as a pair of locality modules $\ontoOne'
\subseteq \O_1$ and $\ontoTwo' \subseteq \O_2$, where $\ontoOne'$ and $\ontoTwo'$ include the semantically related entities to $e_1$ and $e_2$, respectively.
Similarly, the \emph{context} for an alignment $\M$ between two ontologies
$\O_1$ and $\O_2$ is denoted as $\funcalgo{Context}(\M, \O_1, \O_2)=\langle
\ontoOne', \ontoTwo' \rangle$, where  $\ontoOne'$ and $\ontoTwo'$ are modules
including the semantically related entities for the entities $e_1 \in
Sig(\O_1)$ and $e_2 \in Sig(\O_2)$ in each mapping $m=\mappingTwo{e_1}{e_2} \in \M$.
\end{definition}
%
%
Intuitively, as the context of an alignment (\ie $\funcalgo{Context}(\M, \O_1, \O_2)=\langle
\ontoOne', \ontoTwo' \rangle$) semantically characterises the entities involved
in that alignment, a matching task $\MT=\langle \ontoOne, \ontoTwo \rangle$ can be reduced to the task $\MTMO=\langle \ontoOne', \ontoTwo' \rangle$ without information loss in terms of finding $\M$  (\ie $\funcalgo{Coverage}(\MTMO, \M)=\M$).
For example, in the small OAEI \emph{largebio} tasks~\cite{DBLP:conf/semweb/AlgergawyCFFFHH18,DBLP:conf/semweb/AlgergawyFFFHHJ19} systems are given the context of the reference alignment as a (reduced) matching task (\eg  $\MTRAFN = \funcalgo{Context}(\MRAFN,\
\FMA, \NCI)=\langle \FMA', \NCI' \rangle$), instead of the whole FMA and NCI ontologies.

\input{tableIF.tex}

\subsection{Indexing the ontology vocabulary}
\label{sec:lex}

We rely on a semantic inverted index (we will refer to this index as \lex). This index maps sets of words to the 
entities where these words appear.
%
\lex encodes the labels of all entities of the input ontologies $\ontoOne$ and $\ontoTwo$, including their lexical variations (\eg preferred labels, synonyms), in the form of \emph{key-value} pairs
where the key is a set of words and the value is a set of entities
such that the set of words of the key
appears in (one of) the entity labels. 
Similar indexes are commonly  used  in  information  retrieval applications \cite{DBLP:books/daglib/0031897}, Entity Resolution systems \cite{blocking-er-2019}, and also exploited in ontology alignment systems (\eg LogMap~\cite{Jimenez-Ruiz:2011}, ServOMap~\cite{servomap14} 
and AML~\cite{aml18}) to reduce the search space and enhance the matching process.
Table~\ref{table:ext_if} shows a few example entries of \lex for two input ontologies.

\lex is created as follows.
\begin{inparaenum}[\it (i)]
  \item Each label associated to an ontology entity is split into a set of
  words; for example, the label ``Lunate facet of hamate'' is split into the set \{``lunate'', ``facet'', ``of'', ``hamate''\}.
  \item Stop-words  are removed from the set of words. 
  \item Stemming techniques are applied to each word (\ie \{``lunat'', ``facet'', ``hamat''\}). 
  \item Combinations of subsets of words also serve as keys in \lex; for example,
  \{``lunat'', ``facet''\}, \{``hamat'', ``lunat''\} and so on.\footnote{In
  order to avoid a combinatorial blow-up, the number of computed subsets of words
  is limited.}
  \item Entities leading to the same (sub)set of words are associated to the
  same key in \lex, for example \{``disorder''\} is associated with three entities.
  Finally, 
  \item entries in \lex pointing to entities of only one ontology or associated to a number of entities larger than $\alpha$ are not considered.\footnote{In the experiments we used $\alpha=60$.} 
\end{inparaenum}
Note that a single entity label may lead to several entries in \lex, and
each entry in \lex points to one or more entities.


\vspace{-0.1cm}

\subsection{Covering matching subtasks}
\label{sec:cmt}

Each entry (\ie a \emph{key-value} pair) in \lex
is a source of candidate mappings. For
instance, the example in Table \ref{table:ext_if} suggests that there is a
candidate 
mapping $m=\mappingTwo{\cem{\ontoOne \nsp:\nsp
Disorder\_of\_stomach}}{\cem{\ontoTwo \nsp:\nsp Pregnancy\_disorder}}$ 
since 
these entities 
are associated to
the \emph{\{``disorder''\}} entry in \lex.
These mappings are not necessarily correct but will link lexically-related entities, that is, those entities sharing at least one word among their labels (\eg ``disorder'').
Given a subset of entries or rows of \lex (\ie $l \subseteq \lex$), the function 
$\funcalgo{Mappings}(l)=\M^{l}$ provides the set of mappings derived from $l$.  We refer to the
set of all (potential) mappings suggested by \lex (\ie $\funcalgo{Mappings}(\lex)$) as $\MLex$. $\MLex$ represents a manageable subset of the Cartesian product between the entities of the input ontologies. For example, \lex suggest around $2\times10^4$ potential mappings for the matching task $\MTFN =  \langle \FMA, \NCI \rangle$, while the Cartesian product between $\FMA$ and $\NCI$ involves more than $5\times10^9$ mappings.


Since standard ontology alignment systems rarely discover mappings outside
$\MLex$, the context of $\MLex$ (recall Definition \ref{def:context1})
can be seen as a
reduced matching task $\MTLex=\funcalgo{Context}(\MLex, \O_1, \O_2)=\langle \ontoOneP{\lex}, \ontoTwoP{\lex} \rangle$ of the original task $\MT=\langle \ontoOne, \ontoTwo \rangle$. However, the modules $\ontoOneP{\lex}$ and $\ontoTwoP{\lex}$, although smaller than
$\ontoOne$ and $\ontoTwo$, can still be challenging for many ontology matching systems. A solution is to divide or cluster the entries in \lex to lead to several tasks involving smaller ontology modules.

\begin{definition}[Matching subtasks from \lex]
\label{def:partIF}
Let $\MT=\langle \ontoOne, \ontoTwo \rangle$ be a matching task, \emph{\lex} the inverted index of the ontologies $\ontoOne$ and $\ontoTwo$, and $\{l_1,\ldots,l_n\}$ a set of $n$ clusters of entries in \emph{\lex}. 
We denote the set of matching subtasks from \emph{\lex} as
$\PMT^{n} = \{\MTLex_1,\ldots,\MTLex_n\}$ where each cluster $l_i$ leads to the matching subtask $\MTLex_i=\langle \ontoOneP{i},
\ontoTwoP{i} \rangle$, such that $\funcalgo{Mappings}(l_i) = \MLex_i$ is the set of mappings suggested by the \emph{\lex} entries in $l_i$ (\ie \emph{key-value} pairs)
and $\ontoOneP{i}$ and $\ontoTwoP{i}$ represent the context of $\MLex_i$ w.r.t. $\ontoOne$ and $\ontoTwo$.
\end{definition}


\medskip
\noindent
\textbf{Quality of the matching subtasks.} 
The matching subtasks in Definition \ref{def:partIF}  rely on \lex and the notion of context, thus it is expected that the tasks in $\PMT^{n}$ will cover most of the mappings $\MS$  that a matching system can compute, that is $\funcalgo{CoverageRatio}(\PMT^n, \MS)$ will be close to $1.0$.
Furthermore, the use of locality modules to compute the context guarantees the extraction of matching subtasks that are suitable to ontology alignment systems in terms of preservation of the logical properties of the given signature.

Intuitively each cluster of \lex will lead 
to a smaller matching task $\MTLex_i$ (with respect to both $\MTLex$ and $\MT$) in terms of
search space. Hence $\funcalgo{SizeRatio}(\MTLex_i, \MT)$ will be smaller than $1.0$.
%
The overall aggregation of ratios (cf. Equation \ref{eq:sizeeratio}) 
depends on the clustering strategy of the entries in \lex and it is also
expected to be smaller than $1.0$.

Reducing the search space
in each matching subtask $\MTLex_i$ has the potential of enabling 
the evaluation of systems that
cannot cope with the original matching task $\MT$ in a given time-frame or
with (limited) computational resources.

\input{tableTasks}

\subsection{Semantic embeddings}
\label{sec:clustering}

We use a \emph{semantic embedding} approach to identify, given~$n$, a set of clusters of 
entries $\{l_1,\ldots,l_n\}$ from \lex.  As in Definition~\ref{def:partIF}, these clusters 
lead to the set of matching subtasks $\PMT^{n} = \{\MTLex_1,\ldots,\MTLex_n\}$.
The \emph{semantic embeddings} aim at representing into the same (vector) space the features 
about the relationships among words and ontology entities that occur in \lex.
Hence, words and entities that belong to similar semantic contexts will typically have
similar vector representations.





\medskip
\noindent
\textbf{Embedding model.}
Our approach currently relies on the \starspace toolkit\footnote{\starspace:
\url{https://github.com/facebookresearch/StarSpace}} and its neural embedding
model~\cite{wu_2017} to learn \emph{embeddings} for the words and
ontology entities in \lex.
We adopt the 
\emph{TagSpace} \cite{DBLP:conf/emnlp/WestonCA14} training setting of
\starspace. 
%
Applied to our setting, \starspace learns associations between a set of
words (\ie keys in \lex) 
and a set of relevant ontology entities (\ie values in \lex).
%
The \starspace model is trained by assigning a $d$-dimensional vector to
each of the relevant features (\eg  
the individual words and the ontology entities in \lex).
%
Ultimately,
the look-up matrix (the matrix of embeddings - latent vectors) is learned by
minimising the loss function in Equation \ref{eq:loss}. 

\medskip
\begin{scriptsize}
\begin{equation}\label{eq:loss}
\!
\sum_{\substack{(w, e) \in E^+,\\e^- \in E^-}}%
L^{batch} (sim(\bm{v}_w, \bm{v}_e), sim(\bm{v}_w, \bm{v}_{e_1^-}), \ldots,\\
sim(\bm{v}_w, \bm{v}_{e_j^-}))
\end{equation}
\end{scriptsize}
%
%

In this loss function we compare positive samples with negative samples. Hence we need to indicate the generator of positive pairs $(w, e) \in E^+$ (in our setting those are \emph{word-entity} pairs from \lex) and the
generator of negative entries $e^- \in E^-$ 
(in our case we sample from the list of entities in the values of \lex).
\starspace follows the strategy by Mikolov et al.
\cite{mikolov_2013} and selects a random subset of $j$ negative examples
for each batch update.
Note that we tailor the generators to the alignment task by sampling from \lex.
The similarity function
$sim$ 
operates on $d$-dimensional vectors (\eg $\bm{v}_w$, $\bm{v}_e$ and
$\bm{v}_e^-$),
in our case we use the standard dot
product in Euclidean space. 


\medskip
\noindent
\textbf{Clustering strategy.}
The semantic embedding of each entry $\varepsilon=(K, V) \in$ \lex is
calculated by concatenating \begin{inparaenum}[\it (i)]
\item the mean vector representation of the vectors associated to each word in the key $K$, with 
\item the mean vector of the vectors of the ontology entities in the value $V$, 
\end{inparaenum}
as in Equation~\ref{eq:vectors}, where $\oplus$ represents the concatenation of
two vectors, $\bm{v}_w$ and $\bm{v}_e$ represents $d$-dimensional vector embeddings learnt by
\starspace, and $\bm{v}_\varepsilon$ is a ($2*d$)-dimension vector.
\begin{small}
\begin{equation}\label{eq:vectors}
\bm{v}_\varepsilon = \frac{1}{|K|}\sum_{w \in K}\bm{v}_w \oplus
\frac{1}{|V|}\sum_{e \in V}\bm{v}_e
\end{equation}
\end{small}

Based on the embeddings $\bm{v}_\varepsilon$ we then perform standard clustering
with the K-means algorithm to obtain the clusters of \lex entries $\{l_1,\ldots,l_n\}$.
For example, following our approach, in the example of Table \ref{table:ext_if}
entries in rows $1$ and $2$ (respectively $3$ and $4$) would belong to the same
cluster.

\medskip
\noindent
\textbf{Suitability of the embedding model.}
Although we could have followed other embedding strategies, we advocated to learn new entity embeddings with \starspace for the following reasons:
\begin{inparaenum}[\it (i)]
\item ontologies, particularly in the biomedical domain, may bring specialised vocabulary that is not fully covered by precomputed word embeddings;
\item to embed not only words but also concepts of both ontologies; and
\item to obtain embeddings tailored to the ontology alignment task (\ie to learn
similarities among words and concepts dependant on the task).
\end{inparaenum}
\starspace provides the required functionalities to embed the semantics of \lex
and identify accurate clusters. Precise clusters will lead to smaller matching
tasks, and thus, to a reduced global size of the computed division of the
matching task $\PMT^{n}$ (cf. Equation~\ref{eq:sizeeratio}).


%% file: tableIF.tex
\begin{table*}[t!]
\caption{Inverted lexical index \lex. 
For readability,
index values have been split into elements of $\ontoOne$ and $\ontoTwo$. `-'
indicates that the ontology does not contain entities for that  
entry.}\label{table:ext_if}%
\renewcommand{\arraystretch}{1.3}
\vspace{-0.3cm}
\centering
{
\begin{footnotesize}

\begin{tabular}{ll}
\begin{tabular}[t]{|l|l|l|l|}
\hline 

\multirow{2}{*}{\textbf{\#}} &\multirow{2}{*}{\textbf{Index key}} & \multicolumn{2}{c|}{\textbf{Index value}}
\\\cline{3-4} 
& & \textbf{Entities $\ontoOne$} & \textbf{Entities $\ontoTwo$} \\\hline

1 &
$\{$ disorder $\}$ & \ontoOne:Disorder\_of\_pregnancy,  \ontoOne:Disorder\_of\_stomach & \ontoTwo:Pregnancy\_Disorder \\

2 &$\{$ disorder, pregnancy $\}$ & \ontoOne:Disorder\_of\_pregnancy & \ontoTwo:Pregnancy\_Disorder \\

3 & $\{$ carcinoma, basaloid $\}$ & \ontoOne:Basaloid\_carcinoma & \ontoTwo:Basaloid\_Carcinoma, \ontoTwo:Basaloid\_Lung\_Carcinoma \\
4 & $\{$ follicul, thyroid, carcinom $\}$ & \ontoOne:Follicular\_thyroid\_carcinoma  & \ontoTwo:Follicular\_Thyroid\_carcinoma  \\


5 & $\{$ hamate, lunate $\}$ & \ontoOne:Lunate\_facet\_of\_hamate & - \\\hline

\end{tabular}

\end{tabular}

\end{footnotesize}
}
\end{table*}

%% file: tableTasks.tex
\begin{table*}[ht!]
\caption{Matching tasks. AMA: Adult
Mouse  Anatomy. DOID: Human Disease Ontology. FMA: Foundational Model
of Anatomy. HPO: Human Phenotype Ontology.  MP: Mammalian Phenotype. NCI: National
Cancer Institute Thesaurus. NCIA: Anatomy fragment of NCI. ORDO: Orphanet Rare
Disease Ontology. SNOMED CT: Systematized Nomenclature of Medicine -- Clinical
Terms. 
Phenotype ontologies downloaded from BioPortal. For all tracks we use the consensus with vote=3 as system mappings $\MS$. The Phenotype track does not have a gold standard so a consensus alignment with vote=2 is used as reference.
}\label{table:tasks}
\renewcommand{\arraystretch}{1.3}
\centering
{
\begin{footnotesize}
\begin{tabular}{|c|c|c||c|c|c|c|}
\hline
\textbf{OAEI track} & \textbf{Source of $\MRA$} & \textbf{Source of $\MS$} & \textbf{Task}  &
\textbf{Ontology} & \textbf{Version} & \textbf{Size (classes)} \\\hline\hline

\multirow{2}{*}{Anatomy} &  \multirow{2}{*}{Manually
created \cite{mousealignment05}} & \multirow{2}{*}{Consensus (vote=3)} & \multirow{2}{*}{AMA-NCIA} & AMA & v.2007 &
2,744 \\
& & & & NCIA & v.2007 & 3,304 \\\hline\hline

\multirow{3}{*}{Largebio} & \multirow{3}{*}{UMLS-Metathesaurus
\cite{umlsassessment11}} & \multirow{3}{*}{Consensus (vote=3)} & FMA-NCI & FMA & v.2.0 & 78,989 \\
& & & ~FMA-SNOMED~ & NCI & v.08.05d &  66,724 \\
& & & SNOMED-NCI & ~SNOMED CT~ &  v.2009 &  306,591\\\hline\hline

\multirow{4}{*}{Phenotype} & 
\multirow{4}{*}{
Consensus alignment (vote=2) \cite{phenotype2017}
}
& \multirow{4}{*}{Consensus (vote=3)}

& \multirow{2}{*}{HPO-MP} & HPO &
~v.2016~ & 11,786 \\
& & & & MP & v.2016 & 11,721 \\
& & &  \multirow{2}{*}{DOID-ORDO} & DOID & v.2016 & 9,248 \\
& & & & ORDO & v.2016 & 12,936 \\\hline
\end{tabular}
\end{footnotesize}
}
\end{table*}

%% file: evaluation.tex

\section{Evaluation}
\label{sec:eval}

In this section we provide empirical evidence to support the suitability of the proposed method to divide the ontology alignment task.
We rely on the datasets of the Ontology Alignment Evaluation Initiative (OAEI) 
\cite{DBLP:conf/semweb/AlgergawyCFFFHH18,DBLP:conf/semweb/AlgergawyFFFHHJ19}, more specifically, on the matching
tasks provided in the \emph{anatomy},
\emph{largebio}
and \emph{phenotype}
tracks.
Table~\ref{table:tasks} provides an overview of these OAEI tasks and the related ontologies and mapping sets.

The methods have been implemented in Java\footnote{Java codes: \url{https://github.com/ernestojimenezruiz/logmap-matcher}} 
and Python\footnote{Python codes: \url{https://github.com/plumdeq/neuro-onto-part}} (neural embedding strategy), tested on a Ubuntu Laptop with an  
Intel Core i9-8950HK CPU@2.90GHz
and allocating up to $25Gb$ of RAM. 
Datasets, matching subtasks, computed mappings 
and other supporting resources are available in the \emph{Zenodo} repository~\cite{zenodo_material_ecai}.
For all of our experiments we used the following \starspace hyperparameters:
\texttt{-trainMode~0 -similarity dot --epoch 100 --dim 64}.



\input{quality_figures.tex}

\subsection{Adequacy of the division approach}
\label{sect:evalPartitioning}

We have evaluated the adequacy of our division strategy in terms of coverage (as
in Equation \ref{eq:covratio}) and size (as in Equation~\ref{eq:sizeeratio}) of the resulting division~$\PMT^n$ 
for each of the matching task in Table~\ref{table:tasks}.
%

\medskip
\noindent
\textbf{Coverage ratio.}
Figures \ref{fig:coveragesRA} and \ref{fig:coveragesMS} shows the coverage of the different 
divisions~$\PMT^n$ with respect to the reference alignment and system computed
mappings, respectively. 
As system mappings we have used the 
consensus alignment with vote=3, that is, mappings~that have been voted by at
least $3$ systems in the last OAEI~campaigns.
The overall coverage results are encouraging: 
\begin{inparaenum}[\it (i)]
\item the divisions $\PMT^n$ cover over $94\%$ of the reference alignments for all tasks, with the exception of the SNOMED-NCI case where coverage ranges from $0.94$ to $0.90$;
%
%
\item when considering system mappings, the coverage for all divisions is over $0.98$ with the exception of AMA-NCIA, where it ranges from $0.956$ to $0.974$; 
\item increasing the number of divisions $n$ tends to \textit{slightly} decrease the coverage in some of the test cases, 
this is an expected behaviour as the computed divisions include different semantic contexts (\ie locality modules) and some relevant entities may fall out the division; finally
%
\item as shown in \cite{pereira:2017}, the results in terms of coverage of state-of-the-art partitioning methods (\eg \cite{hu:2008,Hamdi:2010}) are very low for the OAEI \emph{largebio} track ($0.76$, $0.59$ and $0.67$ as the best results for FMA-NCI, FMA-SNOMED and SNOMED-NCI, respectively), thus, making the obtained results even more valuable.
\end{inparaenum}

\medskip
\noindent
\textbf{Size ratio.} 
The results in terms of the size (\ie search space) of
the selected divisions $\PMT^n$ are
presented in Figure~\ref{fig:sizes}.
%
The search space is improved with respect to the original $\MT$ for all the cases, getting as low as $5\%$ of the original matching task size for the FMA-NCI and FMA-SNOMED cases. 
The gain in the reduction of the search space gets relatively stable after a
given division size; this result is expected since the context provided by
locality modules ensures modules with the necessary semantically related entities.
%
The scatter plot in Figure~\ref{fig:cloud} visualise the size of the source
modules against the size of the target modules for the FMA-NCI matching subtasks with divisions of size  $n \in \{5, 20, 50, 100\}$. For
instance, the (blue) circles represent points $\big(\lvert Sig(\ontoOne^{i})
\rvert, \lvert Sig(\ontoTwo^{i}) \rvert \big)$ being
$\ontoOne^{i}$ and $\ontoTwo^{i}$ the source and target modules (with $i$=$1$,\ldots,$5$)
in the matching subtasks of~$\PMT^5$. It can be noted that, on average, the size of source and target modules decreases as the size of the division increases. For example, the largest task in ~$\PMT^{20}$ is represented in point $(6754, 9168)$, while the largest task in ~$\PMT^{100}$ is represented in point $(2657, 11842)$.

\input{tableSystems.tex}

\medskip
\noindent
\textbf{Computation times.} The time to compute the divisions of the matching task is
tied to the number of locality modules to extract, which can be
computed in polynomial time relative to the size of
the input ontology~\cite{DBLP:journals/jair/GrauHKS08}.
The creation of \lex does not add an important overhead, while the training of
the neural embedding model ranges from $21s$ in AMA-NCI to $224s$ in SNOMED-NCI.
Overall, for example, the required time to compute the division with $100$ matching subtasks
ranges from $23s$ (AMA-NCIA) to approx. $600s$ (SNOMED-NCI). 


\input{evaluation-systems}

%% file: quality_figures.tex
\begin{figure*}[t]
    \centering
    \begin{subfigure}[b]{0.45\textwidth}
        \centering
        \includegraphics[width=\textwidth]{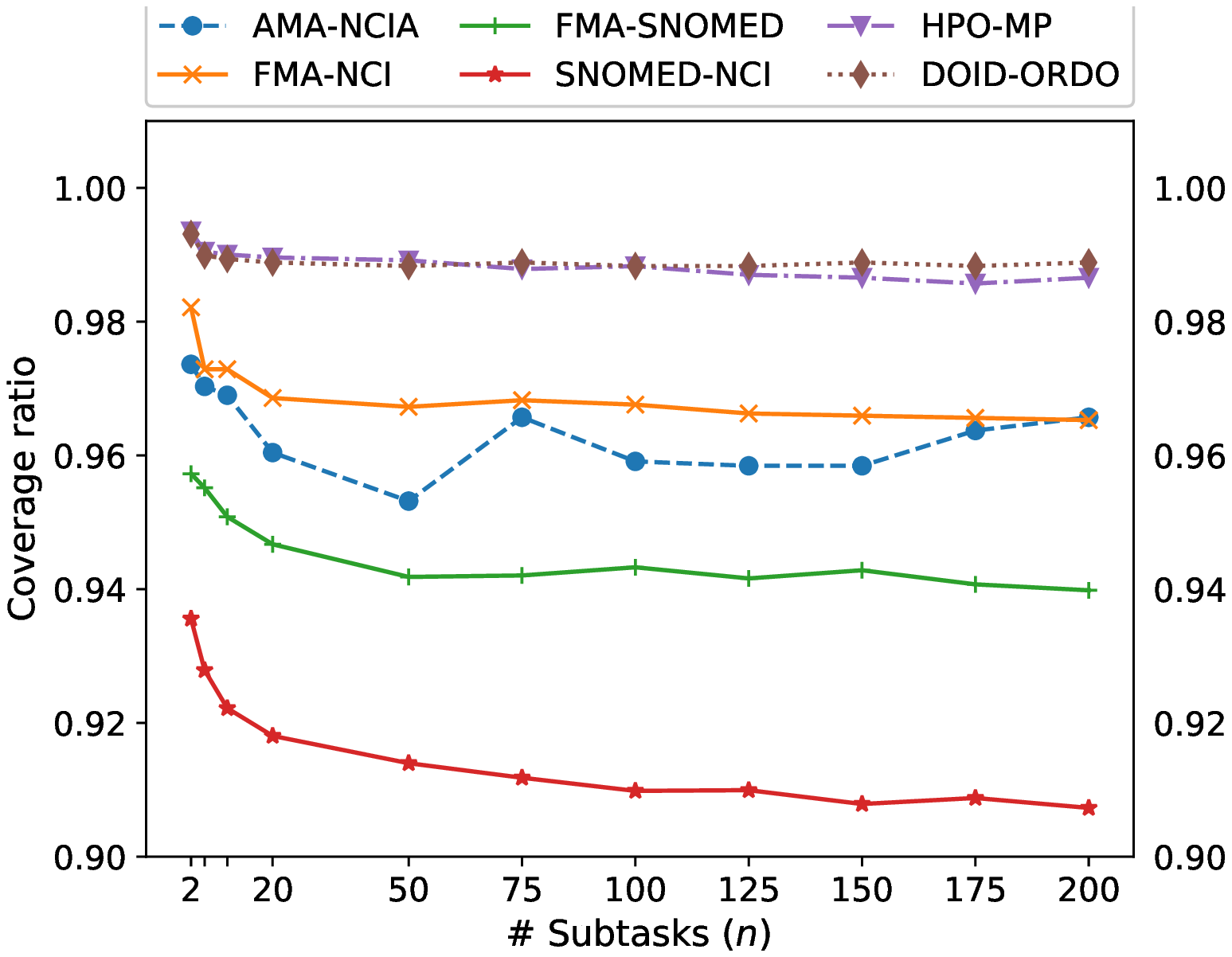}\\[-1ex]
        \caption{$\funcalgo{CoverageRatio}$ of $\PMT^n$  over $\MRA$}
        \label{fig:coveragesRA}
    \end{subfigure}
    \begin{subfigure}[b]{0.45\textwidth}
        \centering
        \includegraphics[width=\textwidth]{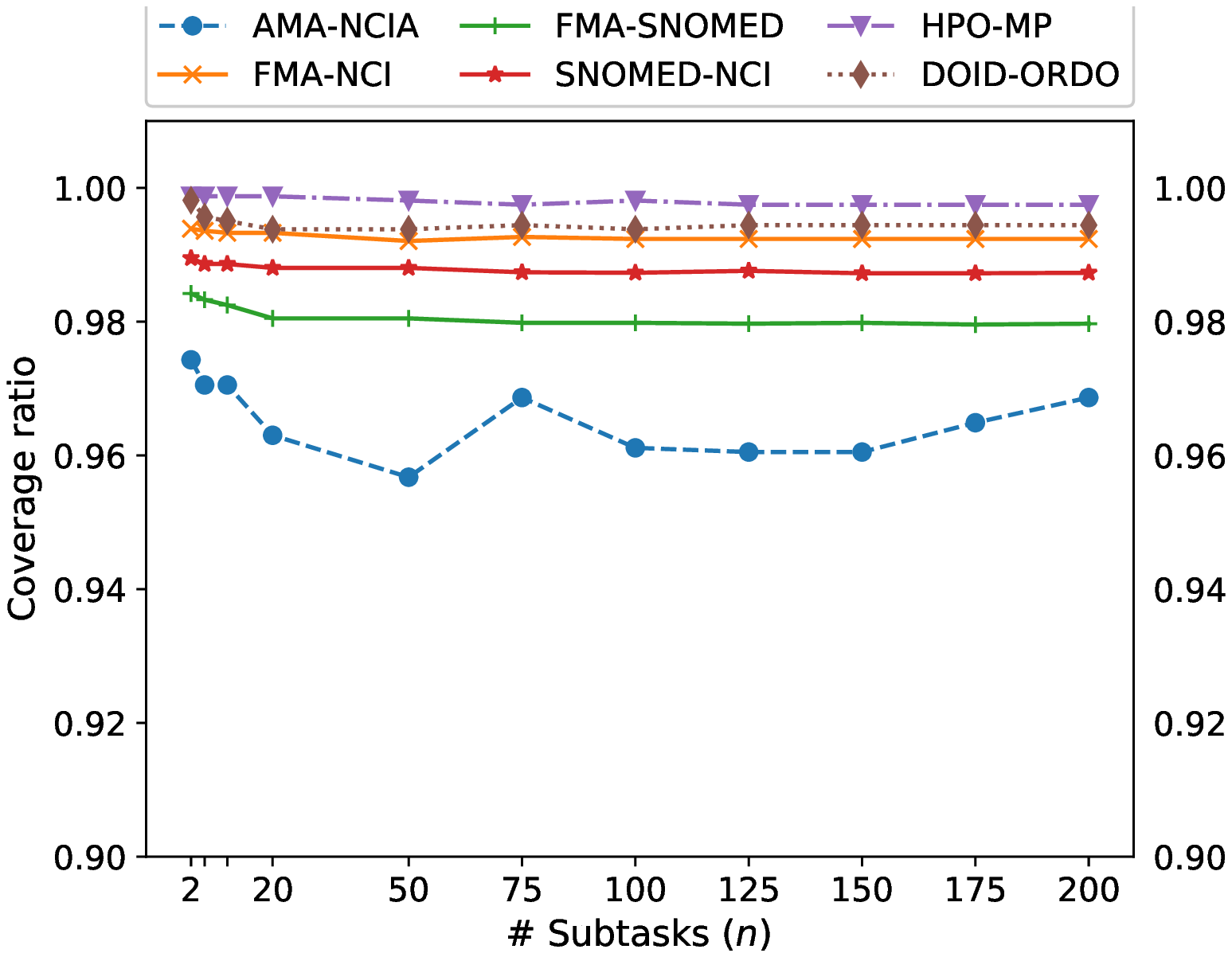}\\[-1ex]
        \caption{$\funcalgo{CoverageRatio}$ of $\PMT^n$ over $\MS$}
        \label{fig:coveragesMS}
    \end{subfigure}
    \\~
    
     \begin{subfigure}[b]{0.45\textwidth}
        \centering
        \includegraphics[width=\textwidth]{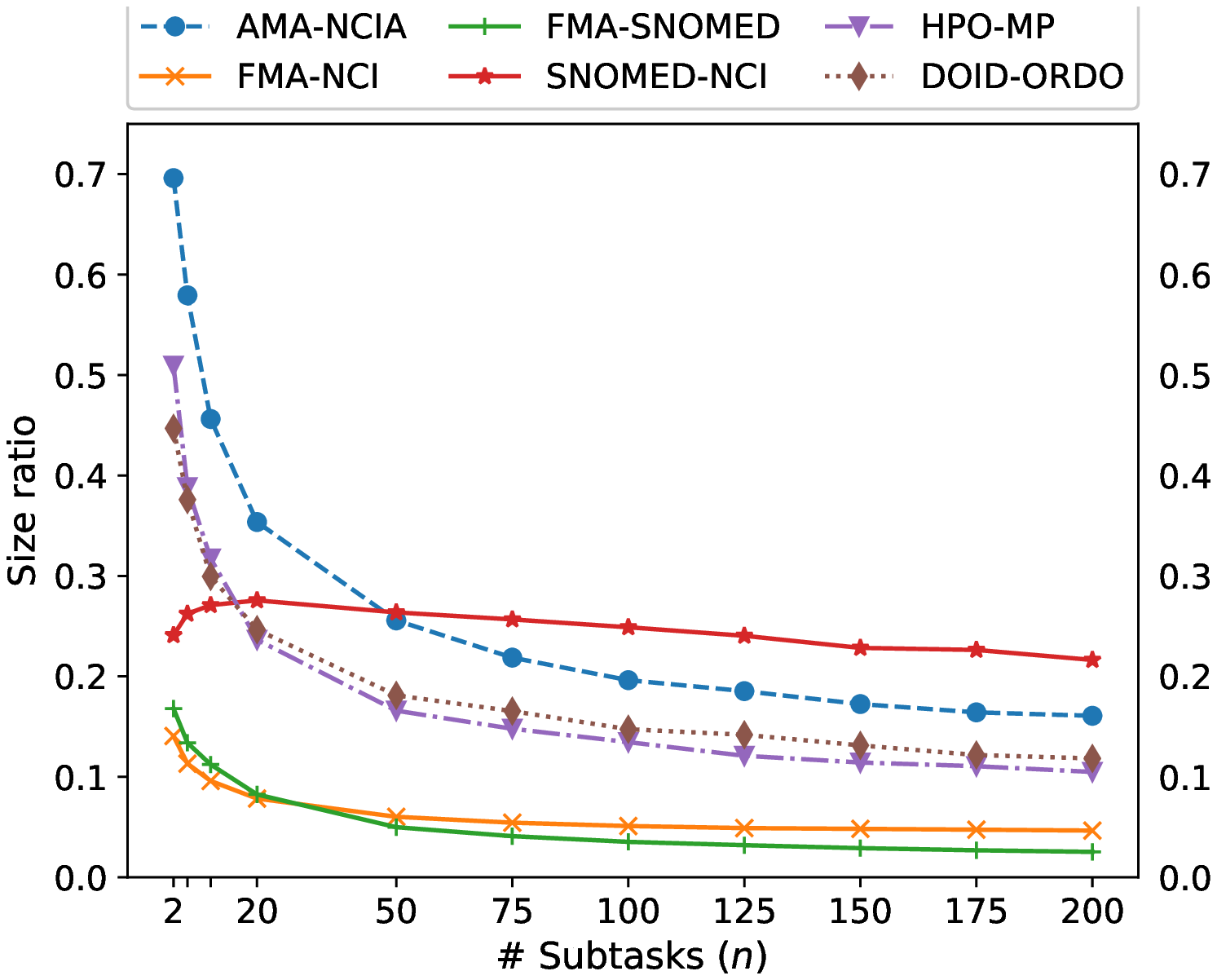}\\[-1ex]
        \caption{$\funcalgo{SizeRatio}$ of $\PMT^n$}
        \label{fig:sizes}
    \end{subfigure}
    \begin{subfigure}[b]{0.45\textwidth}
        \centering
        \includegraphics[width=\textwidth]{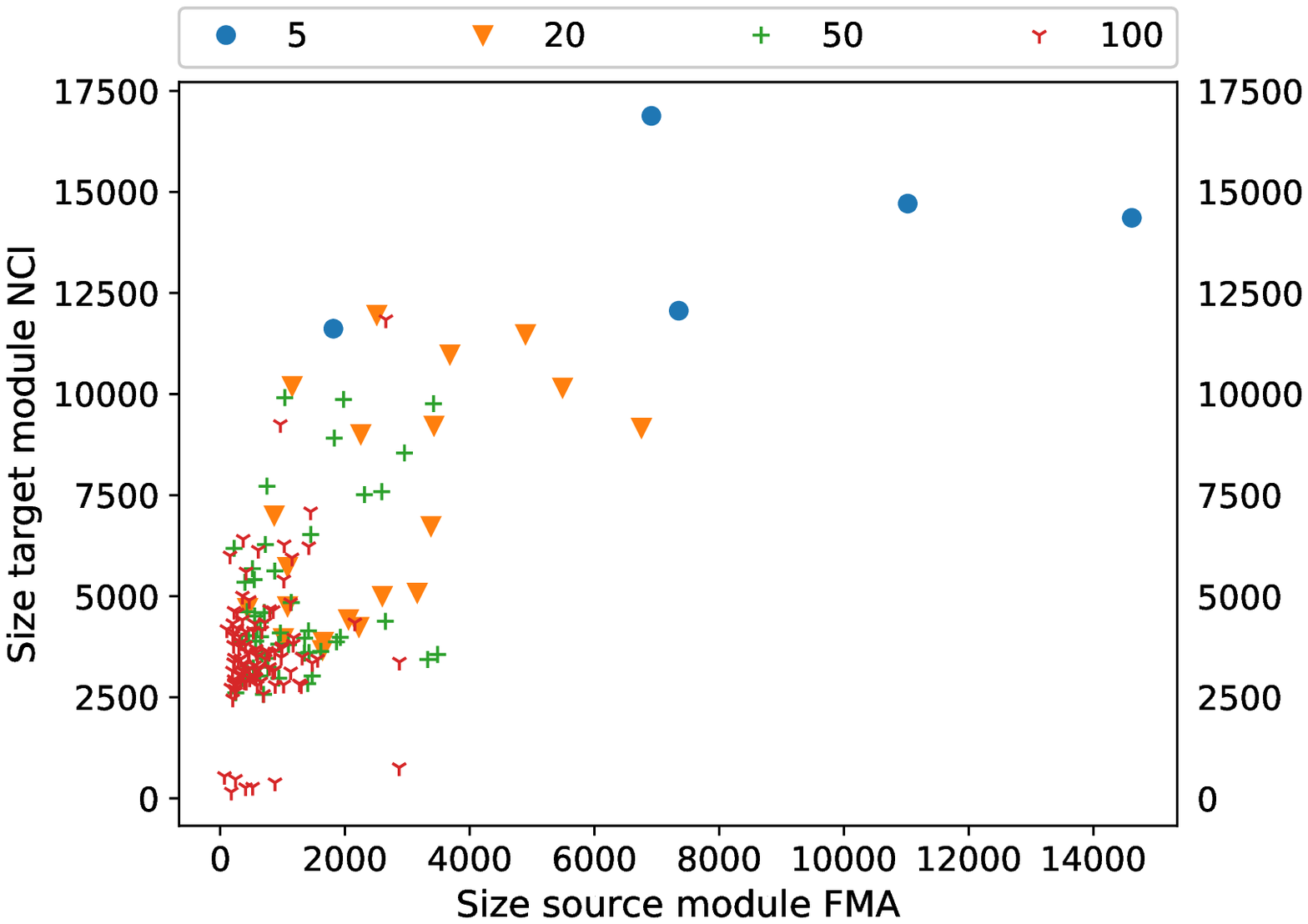}\\[-1ex]
        \caption{Module sizes of $\PMT^n$ for FMA-NCI}
        \label{fig:cloud}
    \end{subfigure}
    
    \caption{Quality measures of $\PMT^n$ with respect to the number of matching subtasks $n$.}
    \label{fig:quality}
\end{figure*}

%% file: tableSystems.tex
\begin{table*}[t!]
\caption{Evaluation of systems that failed to complete OAEI
tasks in the 2015-2018 campaigns. Times reported in
seconds~(s).}\label{table:tools}
\renewcommand{\arraystretch}{1.3}
\centering
{
\begin{footnotesize}
\begin{tabular}{|c|c|c||c|c|c||c|c|c|}
\hline

\multirow{2}{*}{\textbf{Tool}} & \multirow{2}{*}{\textbf{Task}} & 
\textbf{Matching}
& \multicolumn{3}{c||}{\textbf{~~~Performance measures~~~}} &
\multicolumn{3}{c|}{\textbf{Computation times (s)}}\\\cline{4-9}

 & & \textbf{subtasks} &
~\textbf{P}~ &
~\textbf{R}~ &  ~\textbf{F}~ &
~\textbf{Min}~ &  ~\textbf{Max}~ &  ~\textbf{Total}~
\\\hline\hline

\multirow{3}{*}{MAMBA (v.2015)} & \multirow{3}{*}{AMA-NCIA}  & 
5 & 0.870 & \textbf{0.624} & 0.727 & 73 & 785 & 1,981 \\
& & 
10 & ~0.885~ & ~0.623~ & ~0.731~ & 41 & 379 & 1,608 \\
& & 
50 & \textbf{0.897} & 0.623 & \textbf{0.735} & 8 & 154 & ~~~\textbf{1,377}~~~ \\\hline\hline

\multirow{5}{*}{FCA-Map (v.2016)} & 
\multirow{2}{*}{FMA-NCI} & 
20 & \textbf{0.656} & 0.874 & \textbf{0.749} & 39 & 340 & \textbf{2,934}\\
& & 
50 & 0.625 & \textbf{0.875} & 0.729 & 19 & 222 & 3,213 \\\cline{2-9}

& \multirow{2}{*}{FMA-SNOMED} & 
50 & \textbf{0.599} & 0.251 & \textbf{0.354} & 6 & 280 & 3,455\\
& & 
100 & 0.569 & \textbf{0.253} & 0.350 & 5 & 191 & \textbf{3,028} \\\cline{2-9}

& \multirow{2}{*}{SNOMED-NCI} & 
150 & \textbf{0.704} & 0.629 & \textbf{0.664} & 5 & 547 & \textbf{16,822} \\
& & 
200 & 0.696 & \textbf{0.630} & 0.661 & 5 & 395 & 16,874\\\hline\hline

\multirow{4}{*}{SANOM (v.2017) } & 
\multirow{2}{*}{FMA-NCI} & 
20 & \textbf{0.475} & 0.720 & \textbf{0.572} & 40 & 1,467 & 9,374\\
& & 
50 & 0.466 & \textbf{0.726} & 0.568 & 15 & 728 & \textbf{7,069}\\\cline{2-9}

& \multirow{2}{*}{FMA-SNOMED} & 
100 & \textbf{0.145} & \textbf{0.210} & \textbf{0.172} & 3 & 1,044 & 13,073\\
& & 
150 & 0.143 & 0.209 & 0.170 & 3 & 799 & \textbf{10,814}\\\hline\hline

\multirow{4}{*}{POMap++ (v.2018)} & 
\multirow{2}{*}{FMA-NCI} & 
20 & 0.697 & 0.732 & 0.714 & 24 & 850 & 5,448\\
& & 
50 & \textbf{0.701} & \textbf{0.748} & \textbf{0.724} & 11 & 388 & \textbf{4,041}\\\cline{2-9}

& \multirow{2}{*}{FMA-SNOMED} & 
50 & 0.520 & 0.209 & 0.298 & 4 & 439 & 5,879\\
& & 
100 & \textbf{0.522} & \textbf{0.209} & \textbf{0.298} & 3 & 327 & \textbf{4,408}\\\hline\hline

\multirow{4}{*}{ALOD2vec (v.2018)} & 
\multirow{2}{*}{FMA-NCI} & 
20 & 0.697 & 0.813 & 0.751 & 115 & 2,141 & 13,592 \\
& & 
50 & \textbf{0.698} & \textbf{0.813} & \textbf{0.751} & 48 & 933 & \textbf{12,162}\\\cline{2-9}

& \multirow{2}{*}{FMA-SNOMED} & 
100 & 0.702 & 0.183 & 0.29 & 9 & 858 & 12,688\\
& & 
150 & \textbf{0.708} & \textbf{0.183} & \textbf{0.291} & 7 & 581 &
\textbf{10,449}\\\hline
\end{tabular}
\end{footnotesize}
}
\end{table*}

%% file: evaluation-systems.tex
\subsection{Evaluation of OAEI systems}
\label{sect:evalSystems}

In this section we show that the division of the alignment task enables systems that, given
some computational constraints, were unable to complete an OAEI task. 
We have selected the following five systems
from the latest OAEI campaigns, which include novel alignment techniques but failed to scale to very large matching tasks:
MAMBA (v.2015) \cite{mamba2015}, FCA-Map (v.2016) \cite{fcamap2016}, SANOM (v.2017) \cite{sanom17}, ALOD2vec (v.2018) \cite{alod2vec2018} and 
POMap++ (v.2018) \cite{pomap2018}.
%
MAMBA failed to complete the anatomy track, while FCA-Map, SANOM, ALOD2vec and
POMap++ could not complete the largest tasks in the largebio track. 
MAMBA and SANOM threw an out-of-memory exception with $25Gb$, whereas FCA-Map,
ALOD2vec and POMap++ did not complete the tasks within a $6$ hours time-frame.
We have used the SEALS infrastructure to conduct the evaluation
\cite{DBLP:conf/semweb/AlgergawyCFFFHH18,DBLP:conf/semweb/AlgergawyFFFHHJ19}.

Table \ref{table:tools} shows the obtained results 
in terms of  precision, recall, f-measure, and 
computation times (time for the easiest and the hardest task, and total time for all tasks) 
over different divisions $\PMT^n$ computed using our strategy. 
For example, FCA-Map was run over divisions
with 20 and 50 matching subtasks (\ie $n \in \{20, 50\}$) in the FMA-NCI case. 
Note that for each matching subtask a system generates a partial alignment $\MS_i$, the final alignment for the (original) 
matching task is computed as the union of all partial alignments
($\MS=\bigcup_{i=1}^{n} \MS_i$).
The results are encouraging and can be summarised as follows:
\begin{enumerate}[\it i)]
  
  \item We enabled several systems to produce results even for the largest OAEI
  test case (\eg FCA-Map with SNOMED-NCI).
  
  \item The computation times are also very good falling under the $6$ hours time frame, specially given that the (independent) matching subtasks have been run sequentially without 
  parallelization.
  
  \item 
  The size of the divisions, with the exception of FCA-Map, is beneficial in terms of total computation time.
  
  
  \item The increase of number of matching subtasks is positive or neutral for MAMBA, POMap++ and ALOD2vec in terms of f-measure, while it is slightly reduced for FCA-Map and SANOM.
  
  \item Global f-measure results are lower than top OAEI systems; 
  nevertheless, since the above systems could not be evaluated without the
  divisions, these results are obtained without any
  fine-tuning of their parameters. 
  
  \item The computation times of the hardest tasks, as $n$ increases, is also reduced. This has a positive impact in the monitoring of alignment systems as the hardest task is completed in a reasonable time.

\end{enumerate}

%% file: related.tex
\section{Related work}
\label{sec:related}

\smallskip
\noindent
\textbf{Partitioning and blocking.}
Partitioning and modularization 
techniques have been extensively used within the Semantic Web to improve the efficiency when solving the task at hand 
(\eg visualization \cite{stuckenschmidt:2009,agibetov_2015}, 
reuse \cite{DBLP:conf/esws/Jimenez-RuizGSSL08}, debugging
\cite{DBLP:conf/aswc/SuntisrivarapornQJH08}, classification
\cite{DBLP:conf/semweb/RomeroGH12}).
Partitioning or blocking has also been widely used to reduce the complexity of the
ontology alignment task \cite{aml18}. 
In the literature there are two major categories of partitioning
techniques, namely: \emph{independent} and \emph{dependent}. Independent
techniques typically use only the structure of the ontologies and are not concerned about
the ontology alignment task when performing the partitioning. Whereas
dependent partitioning methods rely on both the structure of the ontology and
the ontology alignment task at hand. 
Although our approach does not compute (non-overlapping) partitions of the
ontologies, it can be considered a dependent technique. 

Prominent examples of ontology alignment systems including partitioning
techniques are Falcon-AO \cite{hu:2008}, GOMMA \cite{DBLP:conf/dils/GrossHKR10}, COMA++ \cite{alger:2011} and TaxoMap
\cite{Hamdi:2010}. 
Falcon-AO, GOMMA and COMA++ perform independent partitioning where the clusters of the source and target ontologies are independently
extracted. Then pairs of similar clusters (\ie matching subtasks) are aligned using standard techniques. TaxoMap \cite{Hamdi:2010} 
implements a dependent technique where the partitioning is combined with the matching process. 
TaxoMap proposes two methods, namely: PAP (partition, anchor, partition) and APP (anchor, partition, partition). 
The main difference of these methods is the order of extraction of (preliminary) anchors to discover pairs of 
partitions to be matched (\ie matching subtasks).
SeeCOnt
\cite{Alger:2015} presents a seeding-based clustering technique to
discover
independent clusters in the input ontologies. 
Their approach has been evaluated 
with the Falcon-AO system by replacing its native PBM (Partition-based Block Matching) module~\cite{pbm2016}.
Laadhar et al. \cite{pomap2018} 
have recently integrated within the   
system POMap++ a hierarchical agglomerative clustering
algorithm to divide an ontology into a set of 
partitions.

The above approaches, although presented interesting ideas,
did not provide guarantees about the size and coverage
of the discovered partitions or divisions.
Furthermore, they have not been successfully evaluated on very large ontologies. 
On the one hand, as reported by Pereira et al. \cite{pereira:2017}
the results in terms of coverage of the PBM method of Falcon-OA, and the PAP and APP methods of TaxoMap are very low for the OAEI largebio track. 
On the other hand, as discussed in Section~\ref{sec:eval}, POMap++ fails to scale with the largest largebio~tasks.

Note that the recent work in \cite{DBLP:conf/sac/LaadharGMRTG19} has borrowed from our workshop paper \cite{om18_division} the quality measures presented in Section \ref{sect:quality}. They obtain competitive coverage results for medium size ontologies; however, their approach, as in POMap++, does not scale for large ontologies.

Blocking techniques are also extensively used in Entity Resolution (see \cite{blocking-er-2019} for a survey).
Although related, the problem of blocking in ontologies is different as the logically related axioms for a seed signature play an important role when computing the blocks.



Our dependent approach, unlike traditional partitioning and blocking methods, computes overlapping self-contained modules (\ie locality modules~\cite{DBLP:journals/jair/GrauHKS08}).
Locality modules guarantee the extraction of all semantically related entities
for a given signature. 
This capability enhances the coverage results and enables the inclusion of the
(semantically) relevant information required by an alignment system.
It is worth mentioning that 
the need of self-contained and covering modules, although not thoroughly
studied, was also highlighted in a preliminary work by
Paulheim~\cite{Paulheim:2008}.

\bigskip
\noindent
\textbf{Embedding and clustering.}
Recently, machine learning techniques such as semantic embedding~\cite{cai2018comprehensive} have been investigated for ontology alignment.
They often first learn vector representations of the entities and then predict the alignment \cite{azmy2019matching,zhang2019multi,sun2019transedge}.
However, most of them focus on alignment of ontology individuals (\ie ABox) without considering the ontology concepts and axioms at the terminological level (\ie TBox).
Nkisi-Orji et al. \cite{nkisi2018ontology} predicts the alignment between ontology concepts with Random Forest, but incorporates the embeddings of words alone, without any other semantic components like in our work.
Furthermore, these approaches focus on predicting the alignment, 
while our work aims at boosting an existing alignment system. Our framework could potentially be adopted in systems like in \cite{nkisi2018ontology} if facing scalability problems for large ontologies. 

Another piece of related work is the clustering of semantic components, 
using the canopy clustering algorithm \cite{mccallum2000efficient}
where objects are grouped into canopies and each object can be a member of multiple canopies.
For example, Wu et al. \cite{wu2018towards} first extracted canopies (i.e., mentions) from a knowledge base,
and then grouped the entities accordingly so as to finding out the entities with the same semantics (\ie canonicalization).
As we focus on a totally different task -- ontology alignment,
the context that can be used, such as the embeddings for the words and
ontology entities in \lex, is different from these works, which leads to a different clustering method.

%% file: discussion.tex
\section{Conclusions and future work}
\label{sec:disc}

We have developed a novel framework to split the ontology alignment
task into several matching subtasks based on a semantic inverted index,
locality modules, and a neural embedding model. 
We have performed a comprehensive evaluation which suggests that
the obtained divisions are suitable in practice in terms of both coverage and size. 
The division of the matching task allowed us to obtain results for
five systems which failed to complete these tasks in the past.
We have focused on systems failing to complete a task, but a suitable adoption and integration of the presented framework within the pipeline of any ontology alignment system has the potential to improve the results in terms of
computation times.

\medskip
\noindent
\textbf{Opportunities.}
Reducing the ontology matching task into smaller and more manageable tasks may
also bring opportunities to enhance 
\begin{inparaenum}[\it (i)]
\item user interaction \cite{ker2019}, 
\item reasoning and repair \cite{DBLP:phd/dnb/Meilicke11}, 
\item benchmarking and monitoring \cite{DBLP:conf/semweb/AlgergawyCFFFHH18,DBLP:conf/semweb/AlgergawyFFFHHJ19}, and
\item parallelization.
\end{inparaenum}
%
The computed independent matching subtasks can potentially be run
in parallel in evaluation platforms like the HOBBIT \cite{hobbit16}. The current evaluation was  conducted sequentially as \begin{inparaenum}[\it (i)]
\item the SEALS instance only allows running one task at a time, and
\item the evaluated systems were not designed to run several tasks in parallel; for instance, we managed to run MAMBA outside SEALS, but it relies on a MySQL database and raised a concurrent access exception. 
\end{inparaenum}

\medskip
\noindent
\textbf{Impact on the f-measure.}
As shown in Section \ref{sect:evalSystems}, the impact of the number of divisions on the f-measure depends on the evaluated systems. In the near future we aim at conducting an extensive evaluation of our framework over OAEI systems able to deal with the largest tasks in order to obtain more insights about the impact on the f-measure. In \cite{arxiv18_division} we reported a preliminary evaluation where YAM-Bio \cite{yambio2017} and AML \cite{aml2013} kept similar f-measure values, while LogMap \cite{Jimenez-Ruiz:2011} had a reduction in f-measure, as the number of divisions increased.

%
%


\medskip
\noindent
\textbf{Number of subdivisions.}
Currently our strategy requires the size of the
number of matching subtasks or divisions as input. The (required) matching subtasks may be known before
hand if, for example, the matching tasks are to be run in parallel in a number
of available CPUs. For the cases where the resources are limited or where a
matching system is known to cope with small ontologies, we plan to design an
algorithm to estimate the number of divisions so that
the size of the matching subtasks in the computed divisions is appropriate to the system and resource constraints.
%

\medskip
\noindent
\textbf{Dealing with a limited or large lexicon.}
The construction of \lex
shares a limitation with state-of-the-art systems when the input ontologies are lexically disparate or in different languages.
In such cases, \lex can be enriched with general-purpose lexicons (\eg
WordNet), more specialised background knowledge (\eg UMLS Metathesaurus) or with translated labels using online services~(\eg Google).
On the other hand, a large lexicon may also have an important impact in the computation times. Our conducted evaluation shows, however, that we can cope with very large ontologies with a rich lexicon (\eg NCI Thesaurus).

\medskip
\noindent
\textbf{Notion of context.}
Locality-based modules are typically much smaller than the whole ontology and they have led to very good results in terms of size and coverage. We plan, however,
to study different notions of \emph{context}
of an alignment (\eg the tailored modules proposed in
\cite{DBLP:journals/jair/RomeroKGH16}) to further improve the results in terms of size while keeping the same level of coverage.


%% file: ecai2020_division.bbl
\begin{thebibliography}{10}

\bibitem{agibetov_2015}
Asan Agibetov, Giuseppe Patan\`e, and Michela Spagnuolo, `Grontocrawler:
  {Graph}-{Based} {Ontology} {Exploration}', in {\em STAG}, (2015).

\bibitem{Alger:2015}
Alsayed Algergawy, Samira Babalou, Mohammad~J. Kargar, and S.~Hashem
  Davarpanah, `{SeeCOnt: A New Seeding-Based Clustering Approach for Ontology
  Matching}', in {\em ADBIS}, (2015).

\bibitem{DBLP:conf/semweb/AlgergawyCFFFHH18}
Alsayed Algergawy et~al., `{Results of the Ontology Alignment Evaluation
  Initiative 2018}', in {\em 13th Int'l Workshop on Ontology Matching}, (2018).

\bibitem{DBLP:conf/semweb/AlgergawyFFFHHJ19}
Alsayed Algergawy et~al., `{Results of the Ontology Alignment Evaluation
  Initiative 2019}', in {\em Int'l Workshop on Ontology Matching}, (2019).

\bibitem{alger:2011}
Alsayed Algergawy, Sabine Massmann, and Erhard Rahm, `A clustering-based
  approach for large-scale ontology matching', in {\em ADBIS}, pp. 415--428,
  (2011).

\bibitem{yambio2017}
Amina Annane, Zohra Bellahsene, Fai{\c{c}}al Azouaou, and Cl{\'{e}}ment
  Jonquet, `{YAM-BIO:} results for {OAEI} 2017', in {\em 12th Int'l Workshop on
  Ontology Matching}, (2017).

\bibitem{DBLP:conf/semweb/RomeroGH12}
Ana {Armas Romero}, Bernardo {Cuenca Grau}, and Ian Horrocks, `{MORe: Modular
  Combination of {OWL} Reasoners for Ontology Classification}', in {\em Int'l
  Sem. Web Conf.}, (2012).

\bibitem{DBLP:journals/jair/RomeroKGH16}
Ana {Armas Romero}, Mark Kaminski, Bernardo {Cuenca Grau}, and Ian Horrocks,
  `{Module Extraction in Expressive Ontology Languages via Datalog Reasoning}',
  {\em J. Artif. Intell. Res.}, {\bf 55}, (2016).

\bibitem{azmy2019matching}
Michael Azmy, Peng Shi, Jimmy Lin, and Ihab~F Ilyas, `Matching entities across
  different knowledge graphs with graph embeddings', {\em CoRR}, {\bf
  abs/1903.06607}, (2019).

\bibitem{mousealignment05}
Olivier Bodenreider, Terry~F. Hayamizu, Martin Ringwald, Sherri de~Coronado,
  and Songmao Zhang, `Of mice and men: Aligning mouse and human anatomies', in
  {\em AMIA Symposium}, (2005).

\bibitem{DBLP:books/daglib/0031897}
Stefan B{\"{u}}ttcher, Charles L.~A. Clarke, and Gordon~V. Cormack, {\em
  {Information Retrieval - Implementing and Evaluating Search Engines}}, {MIT}
  Press, 2010.

\bibitem{cai2018comprehensive}
Hongyun Cai, Vincent~W Zheng, and Kevin Chen-Chuan Chang, `A comprehensive
  survey of graph embedding: Problems, techniques, and applications', {\em IEEE
  Trans. on Know. and Data Eng.}, {\bf 30}(9), (2018).

\bibitem{DBLP:journals/jair/GrauHKS08}
Bernardo {Cuenca Grau}, Ian Horrocks, Yevgeny Kazakov, and Ulrike Sattler,
  `Modular reuse of ontologies: Theory and practice', {\em J. Artif. Intell.
  Res.}, {\bf 31}, (2008).

\bibitem{servomap14}
Gayo Diallo, `An effective method of large scale ontology matching', {\em J.
  Biomedical Semantics}, {\bf 5}, ~44, (2014).

\bibitem{DBLP:books/daglib/0032976}
J{\'{e}}r{\^{o}}me Euzenat and Pavel Shvaiko, {\em Ontology Matching, Second
  Edition}, Springer, 2013.

\bibitem{aml18}
Daniel Faria, Catia Pesquita, Isabela Mott, Catarina Martins, Francisco~M.
  Couto, and Isabel~F. Cruz, `Tackling the challenges of matching biomedical
  ontologies', {\em J. Biomedical Semantics}, {\bf 9}(1), (2018).

\bibitem{aml2013}
Daniel Faria, Catia Pesquita, Emanuel Santos, Matteo Palmonari, Isabel~F. Cruz,
  and Francisco~M. Couto, `{The AgreementMakerLight Ontology Matching System}',
  in {\em {OTM-ODBASE} Conference}, (2013).

\bibitem{OWL2}
Bernardo~Cuenca Grau, Ian Horrocks, Boris Motik, Bijan Parsia, Peter~F.
  Patel{-}Schneider, and Ulrike Sattler, `{OWL 2}: The next step for {OWL}',
  {\em J. Web Semantics}, {\bf 6}(4), (2008).

\bibitem{DBLP:conf/dils/GrossHKR10}
Anika Gro{\ss}, Michael Hartung, Toralf Kirsten, and Erhard Rahm, `On matching
  large life science ontologies in parallel', in {\em Data Integration in the
  Life Sciences (DILS)}, (2010).

\bibitem{Hamdi:2010}
Fay{\c{c}}al Hamdi, Brigitte Safar, Chantal Reynaud, and Ha{\"{i}}fa
  Zargayouna, `Alignment-based partitioning of large-scale ontologies', in {\em
  Advances in Knowledge Discovery and Management}, (2009).

\bibitem{phenotype2017}
Ian Harrow, Ernesto Jim{\'{e}}nez{-}Ruiz, et~al., `Matching disease and
  phenotype ontologies in the ontology alignment evaluation initiative', {\em
  J. Biomedical Semantics}, {\bf 8}(1), (2017).

\bibitem{hu:2008}
Wei Hu, Yuzhong Qu, and Gong Cheng, `Matching large ontologies: A
  divide-and-conquer approach', {\em Data Knowl. Eng.}, {\bf 67},  140--160,
  (2008).

\bibitem{pbm2016}
Wei Hu, Yuanyuan Zhao, and Yuzhong Qu, `{Partition-Based Block Matching of
  Large Class Hierarchies}', in {\em Asian Sem. Web Conf.}, (2006).

\bibitem{zenodo_material_ecai}
Ernesto Jim{\'{e}}nez{-}Ruiz, Asan Agibetov, Jiaoyan Chen, Matthias Samwald,
  and Valerie Cross.
\newblock {Dividing the Ontology Alignment Task [Data set]}, 2019.
\newblock Zenodo. \url{https://doi.org/10.5281/zenodo.3547888}.

\bibitem{arxiv18_division}
Ernesto Jim{\'{e}}nez{-}Ruiz, Asan Agibetov, Matthias Samwald, and Valerie
  Cross, `Breaking-down the ontology alignment task with a lexical index and
  neural embeddings', {\em CoRR}, {\bf abs/1805.12402}, (2018).

\bibitem{om18_division}
Ernesto Jim{\'{e}}nez{-}Ruiz, Asan Agibetov, Matthias Samwald, and Valerie
  Cross, `We divide, you conquer: from large-scale ontology alignment to
  manageable subtasks with a lexical index and neural embeddings', in {\em 13th
  Int'l Workshop on Ontology Matching}, (2018).

\bibitem{Jimenez-Ruiz:2011}
Ernesto Jim{\'{e}}nez{-}Ruiz and Bernardo~Cuenca Grau, `{LogMap: Logic-Based
  and Scalable Ontology Matching}', in {\em Int'l Sem. Web Conf.}, (2011).

\bibitem{umlsassessment11}
Ernesto Jim{\'{e}}nez{-}Ruiz, Bernardo~Cuenca Grau, Ian Horrocks, and
  Rafael~Berlanga Llavori, `Logic-based assessment of the compatibility of
  {UMLS} ontology sources', {\em J. Biomedical Semantics}, {\bf 2}, (2011).

\bibitem{DBLP:conf/esws/Jimenez-RuizGSSL08}
Ernesto Jim{\'{e}}nez{-}Ruiz, Bernardo~Cuenca Grau, Ulrike Sattler, Thomas
  Schneider, and Rafael Berlanga, `{Safe and Economic Re-Use of Ontologies: A
  Logic-Based Methodology and Tool Support}', in {\em European Sem. Web Conf.},
  (2008).

\bibitem{pomap2018}
Amir Laadhar, Fa{\"{i}}za Ghozzi, Imen Megdiche, Franck Ravat, Olivier Teste,
  and Fa{\"{i}}ez Gargouri, `{OAEI} 2018 results of {POMap++}', in {\em 13th
  Int'l Workshop on Ontology Matching}, (2018).

\bibitem{DBLP:conf/sac/LaadharGMRTG19}
Amir Laadhar, Fa{\"{i}}za Ghozzi, Imen Megdiche, Franck Ravat, Olivier Teste,
  and Fa{\"{i}}ez Gargouri, `Partitioning and local matching learning of large
  biomedical ontologies', in {\em 34th {ACM/SIGAPP} Symposium on Applied
  Computing {SAC}}, (2019).

\bibitem{ker2019}
Huanyu Li, Zlatan Dragisic, Daniel Faria, Valentina Ivanova, Ernesto
  Jiménez-Ruiz, Patrick Lambrix, and Catia Pesquita, `User validation in
  ontology alignment: functional assessment and impact', {\em The Knowledge
  Engineering Review}, {\bf 34}, (2019).

\bibitem{mccallum2000efficient}
Andrew McCallum, Kamal Nigam, and Lyle~H Ungar, `Efficient clustering of
  high-dimensional data sets with application to reference matching', in {\em
  6th ACM SIGKDD}, (2000).

\bibitem{DBLP:phd/dnb/Meilicke11}
Christian Meilicke, {\em Alignment incoherence in ontology matching}, Ph.D.\
  dissertation, University of Mannheim, 2011.

\bibitem{mamba2015}
Christian Meilicke, `{MAMBA} - results for the {OAEI} 2015', in {\em 10th Int'l
  Workshop on Ontology Matching}, (2015).

\bibitem{mikolov_2013}
Tomas Mikolov, Ilya Sutskever, Kai Chen, Greg Corrado, and Jeffrey Dean,
  `Distributed representations of words and phrases and their
  compositionality', {\em arXiv}, (oct 2013).

\bibitem{sanom17}
Majid Mohammadi, Amir~Ahooye Atashin, Wout Hofman, and Yao{-}Hua Tan, `{SANOM}
  results for {OAEI} 2017', in {\em 12th Int'l Workshop on Ontology Matching},
  (2017).

\bibitem{hobbit16}
Axel{-}Cyrille~Ngonga Ngomo, Alejandra Garc{\'{i}}a{-}Rojas, and Irini
  Fundulaki, `{HOBBIT:} holistic benchmarking of big linked data', {\em {ERCIM}
  News}, {\bf 2016}(105), (2016).

\bibitem{nkisi2018ontology}
Ikechukwu Nkisi-Orji, Nirmalie Wiratunga, Stewart Massie, Kit-Ying Hui, and
  Rachel Heaven, `Ontology alignment based on word embedding and random forest
  classification', in {\em ECML/PKDD}, (2018).

\bibitem{blocking-er-2019}
George Papadakis, Dimitrios Skoutas, Emmanouil Thanos, and Themis Palpanas, `{A
  Survey of Blocking and Filtering Techniques for Entity Resolution}', {\em
  CoRR}, {\bf abs/1905.06167}, (2019).

\bibitem{Paulheim:2008}
Heiko Paulheim, `{On Applying Matching Tools to Large-scale Ontologies}', in
  {\em 3rd Int'l Workshop on Ontology Matching}, (2008).

\bibitem{pereira:2017}
Sunny Pereira, Valerie Cross, and Ernesto Jim{\'{e}}nez{-}Ruiz, `On
  partitioning for ontology alignment', in {\em Int'l Sem. Web Conf. (Poster)},
  (2017).

\bibitem{alod2vec2018}
Jan Portisch and Heiko Paulheim, `{ALOD2Vec matcher}', in {\em 13th Int'l
  Workshop on Ontology Matching}, (2018).

\bibitem{DBLP:journals/tkde/ShvaikoE13}
Pavel Shvaiko and J{\'{e}}r{\^{o}}me Euzenat, `Ontology matching: State of the
  art and future challenges', {\em {IEEE} Trans. Knowl. Data Eng.}, {\bf
  25}(1), (2013).

\bibitem{stuckenschmidt:2009}
Heiner Stuckenschmidt and Anne Schlicht, `Structure-based partitioning of large
  ontologies', in {\em Modular Ontologies: Concepts, Theories and Techniques
  for Knowledge Modularization}, Springer, (2009).

\bibitem{sun2019transedge}
Zequn Sun, Jiacheng Huang, Wei Hu, Muhao Chen, Lingbing Guo, and Yuzhong Qu,
  `{TransEdge: Translating Relation-Contextualized Embeddings for Knowledge
  Graphs}', in {\em Int'l Sem. Web Conf. (ISWC)}, (2019).

\bibitem{DBLP:conf/aswc/SuntisrivarapornQJH08}
Boontawee Suntisrivaraporn, Guilin Qi, Qiu Ji, and Peter Haase, `A
  modularization-based approach to finding all justifications for {OWL} {DL}
  entailments', in {\em Asian Sem. Web Conf.}, (2008).

\bibitem{DBLP:conf/emnlp/WestonCA14}
Jason Weston, Sumit Chopra, and Keith Adams, `{\#}tagspace: Semantic embeddings
  from hashtags', in {\em Conference on Empirical Methods in Natural Language
  Processing ({EMNLP})}, (2014).

\bibitem{wu_2017}
Ledell Wu, Adam Fisch, Sumit Chopra, Keith Adams, Antoine Bordes, and Jason
  Weston, `{StarSpace: Embed All The Things!}', {\em arXiv}, (2017).

\bibitem{wu2018towards}
Tien-Hsuan Wu, Zhiyong Wu, Ben Kao, and Pengcheng Yin, `Towards practical open
  knowledge base canonicalization', in {\em 27th ACM Int'l Conf. on Inf. and
  Knowledge Management}, (2018).

\bibitem{zhang2019multi}
Qingheng Zhang, Zequn Sun, Wei Hu, Muhao Chen, Lingbing Guo, and Yuzhong Qu,
  `Multi-view knowledge graph embedding for entity alignment', in {\em 28th
  Int'l Joint Conf. on Art. Intell. (IJCAI)}, (2019).

\bibitem{fcamap2016}
Mengyi Zhao and Songmao Zhang, `{FCA-Map} results for {OAEI} 2016', in {\em
  11th Int'l Workshop on Ontology Matching}, (2016).

\end{thebibliography}
